\documentclass{article}

    \PassOptionsToPackage{numbers}{natbib}

    \usepackage[preprint]{neurips_2021}

\usepackage[utf8]{inputenc} 
\usepackage[T1]{fontenc}    
\usepackage{hyperref}       
\usepackage{url}            
\usepackage{booktabs}       
\usepackage{amsfonts}       
\usepackage{nicefrac}       
\usepackage{microtype}      
\usepackage{xcolor}         
\usepackage{amsmath}        

\usepackage{graphicx}
\graphicspath{ {./figures/} }

\newif\ifcomments
\commentstrue
\ifcomments
\usepackage[normalem]{ulem}
\definecolor{CMpurple}{rgb}{0.6,0.18,0.64}
\newcommand\cm[1]{\textcolor{CMpurple}{\textsf{\scriptsize[\textbf{CM\@:} #1]}}}
\newcommand\cmi[1]{\textcolor{CMpurple}{#1}}
\newcommand\cmm[1]{\marginpar{\raggedright\tiny\textcolor{CMpurple}{\textsf{{\bfseries CM\@:} #1}}}}
\newcommand\cms{\bgroup\markoverwith{\textcolor{CMpurple}{\rule[.4ex]{2pt}{0.8pt}}}\ULon}
\else
\newcommand\cm[1]{}
\newcommand\cmi[1]{\ignorespaces}
\newcommand\cmm[1]{}
\newcommand\cms[1]{#1}
\fi

\title{Neural \textit{Abstructions}: Abstractions that Support Construction for Grounded Language Learning}

\author{%
  Kaylee Burns
  \thanks{Correspond to: \texttt{kayburns@stanford.edu}}
    \\
  Stanford University\\
   \And
   Christopher D. Manning \\
   Stanford University \\
   \AND
   Li Fei-Fei \\
   Stanford University \\
}

\begin{document}

\maketitle

\begin{abstract}
Although virtual agents are increasingly situated in environments where natural language is the most effective mode of interaction with humans, these exchanges are rarely used as an opportunity for learning.
%
Leveraging language interactions effectively requires addressing limitations in the two most common approaches to language grounding: semantic parsers built on top of fixed object categories are precise but inflexible and end-to-end models are maximally expressive, but fickle and opaque.
Our goal is to develop a system that balances the strengths of each approach so that users can teach agents new instructions that generalize broadly from a single example.
%
We introduce the idea of neural \textit{abstructions}: a set of constraints on the inference procedure of a label-conditioned generative model that can affect the meaning of the label in context.
Starting from a core programming language that operates over abstructions, users can define increasingly complex mappings from natural language to actions.
We show that with this method a user population is able to build a semantic parser for an open-ended house modification task in Minecraft.
The semantic parser that results is both flexible and expressive: the percentage of utterances sourced from redefinitions increases steadily over the course of 191 total exchanges, achieving a final value of 28\%.
\end{abstract}
\section{Introduction}
\label{sec:intro}
As language learning agents become embodied in virtual and physical worlds that are shared with users, they are presented with the opportunity to curate rich data from humans for little to no cost.
    For example, when an agent misunderstands something, it can simply ask the human for input and guidance.
    Humans can explain unfamiliar concepts and describe procedures for accomplishing new tasks.
    Making these exchanges frictionless---and perhaps even beneficial to the user---incentivizes the human and agent to collaboratively construct rich mappings from natural language to actions or programs.
However, the current machine learning toolkit needs stronger solutions for learning from complex instructions quickly and robustly.
    The goal of this work is to develop tools that allow users to define new instructions for a natural language system that can be adopted immediately and generally.

\citet{wang-etal-2017-naturalizing} first demonstrated how a user population can collaboratively create a more natural interface into a set of programs.
The authors create a programming language for the task of building structures in voxel-based environment. Users then \textit{naturalize} this programming language by providing pairings between natural language requests and programs.
However, this model is very brittle as there is no room for users to specify distributions over object categories. For example, a reference to a tree can only ever refer to one instantiation of ``tree''.
Perhaps,
we could enumerate all possible reference objects and develop generative or discriminative models to stand in for those references \citep{karamcheti2020learning}, but these approaches ignore a whole class of creative design problems where fixed object categories are not expressive enough to capture all potential user commands.
Another alternative is to build end-to-end models that directly map language to a given modality with no modularity governing the interface between them \citep{lu2019vilbert,rxr,Ramesh2021ZeroShotTG},
but these approaches suffer from the opacity and brittleness for which deep learning is infamous. Maintaining reliability helps users have a consistent and enjoyable experience with a language-equipped agent.

\begin{figure}[!t]
    \center{\includegraphics[width=\textwidth]{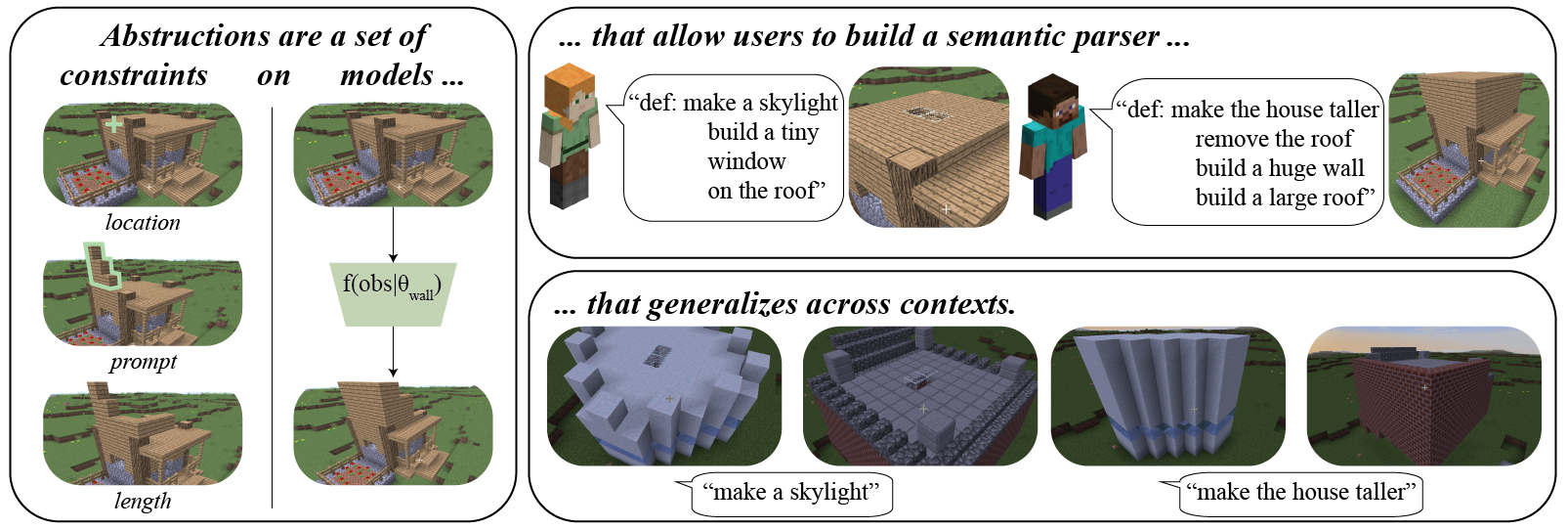}}
    \caption{\label{fig:pull} (left) Abstructions define a set of generative models and associated constraints so that they can be repurposed to represent arbitrarily complex reference objects. (right) Abstructions enable single-shot transfer of user-defined commands, enabling the emergence of a semantic parser that develops precision and flexibility in a target environment.} 
\end{figure}

We introduce the concept of \textit{abstructions}, a characterization of the generative models and user control inputs that enable the emergence of a flexible yet precise semantic parser from interactions with a population of users. Instead of beginning with a collection of fixed object categories, our parser binds language to a set of label-conditioned generative models. The parser is also made aware of constraints that affect the quality, performance, and meaning of these generative models, such as the initial location, build prompt, and length of generation. Adding these constraints to the generation process can change the meaning of a given model's output in context, as illustrated in Figure \ref{fig:pull}. Our blend of generative models and sets of constraints on them allows us to build a semantic parser that is sensitive to context while being able to adapt robustly to category knowledge given to us by users.

We evaluate the ability of a user population to develop a precise yet flexible semantic parser from abstructions in a creative building task in Minecraft using the CraftAssist \citep{gray2019craftassist} framework. Specifically, users are placed in a Minecraft session with our language agent and are instructed to make any desired modifications to a given house by talking directly with the agent. We show that over time, users rely increasingly on induced utterances and that the number of failed parses decreases dramatically. We demonstrate how newly defined commands can be applied to a wide variety of homes, immediately after being defined.
Collectively, these results show that abstructions have the potential to create a language interface that deftly blends precision and flexibility unlike past approaches to language grounding.

\section{Related Work}
\label{sec:related}





\paragraph{Vision and Language.} Static, supervised datasets have served as the foundation for bridging language to various modalities. A rich body of work focuses on the development of machine learning models that can successfully describe, reason about, and navigate within the visual world.
%
In developing solutions \citep{Yu_2016_CVPR,fukui-etal-2016-multimodal,gao2017videocw,Anderson_2018_CVPR,fried2018speaker,perez2018film,lu2019vilbert,Su2020VL-BERT,radford2021clip,ramesh2021zeroshot} for language and video descriptions \citep{Lin2014MicrosoftCC,sharma2018conceptual,krishna2017dense} and visual question answering \citep{VQA,hudson2018gqa,embodiedqa}, researchers have identified critical limitations in the modern toolkit for multi-modal learning.
Significant ablations don't result in significant performance drops \citep{Devlin2015ExploringNN,Kojima2020WhatIL}, vision is ``ignored'' in favor of language cues that are well correlated with prediction \citep{objectHallucination,thomason-etal-2019-shifting}, and models overfit to spurious correlations \citep{cirik-etal-2018-visual} that undermine generalization \citep{agrawal-etal-2016-analyzing}. Many of these concerns have been addressed by curating balanced datasets \citep{Agrawal2017CVQAAC,johnson2017clevr,Suhr2017ACO,Goyal2018MakingTV,Suhr2019ACF}, introducing auxiliary losses that counteract spurious correlations \citep{Ramakrishnan2018OvercomingLP}, and designing models with modularity in mind \citep{Andreas2016NeuralMN,yi2018nsvqa,Mao2019NeuroSymbolic,Hu_2018_ECCV}. But, disembodied from the environment in which the data was collected, agents are deprived of rich interactions to further structure their learning.

\paragraph{Structuring Other Modalities with Language.} Language data can provide structure and insight for tasks that don't require it explicitly. Leveraging the compositionality of language by including it as an additional training input can improve classification performance in fine-grained \citep{he2017finegrainedic} or few-shot \citep{mu2020shaping} settings and can improve exploration when used to set goals intrinsically for reinforcement learning \citep{colas2020languageaa}. Language can also provide explanations for visual classification decisions, which gives practitioners insight into spurious correlations learned by visual models \citep{hendricks2016generating,Hendricks2018GroundingVE,Mu2020CompositionalEO}. Most similar to our contribution is work that uses language feedback at training \citep{Goyal2019UsingNL,sumers2021learningrf,srivastava-etal-2017-joint} or inference \citep{rupprecht2018guide} time to guide and improve prediction results. However, instead of using language as an additional input at training or inference, we use natural language commands in combination with other user inputs to predict a set of constraints that affects the inference procedure of a generative model.

\paragraph{Learning Actions from Commands.} The goal of our work is to learn a mapping from natural language instructions to actions. Some general frameworks for mapping instructions to actions include language-conditioned reinforcement learning \citep{chaplot2017gated,Bahdanau2019LearningTU,Blukis2019LearningTM}, semantic parsers learned from supervision \citep{guu-etal-2017-language,srinet-etal-2020-craftassist}, and a supervised mapping from instruction to action \citep{misra-etal-2018-mapping}. Tasks that focus on this problem include instruction guided navigation \citep{Anderson2018VisionandLanguageNI,rxr,misra-etal-2018-mapping} and cooperative localization \citep{hahn_embodiedlocalization_2020}. In contrast with these tasks, we focus on an open-ended creative design task like \citet{kim-etal-2019-codraw} where agents and humans can take actions collaboratively like \citet{Suhr2019cerealbar}.

\paragraph{Language Learning from Interactions.} Leveraging interactions with humans in the interest of improved learning outcomes has been studied in a variety of settings. Interactive dialogue can be used to bootstrap the capabilities of a semantic parser \citep{artzi-zettlemoyer-2011-bootstrapping,Thomason2015LearningTI}, learn concepts from single examples \citep{Zhang2018InteractiveLA}, narrow-down classification decisions during inference \citep{Yu2020interactiveclass} or training \citep{viscur_corl_2018}, or improve visual concept models in an online fashion \citep{thomason2019improving}. Interactions need not be restricted to language: agents can also infer programs directly from examples \citep{Nye2020RepresentingPP}. Notably, \citet{shah2021basalt} defines tasks for learning from human interactions in Minecraft. In our work, we use interactions to define new commands. Our goal is to naturalize a programming language through user-provided redefinitions, as described in \citet{wang-etal-2017-naturalizing}. We leverage a modernization of this approach presented in \citet{karamcheti2020learning}. However, unlike in these works, abstructions allow users to quickly define new object categories that are context sensitive, enabling naturalization in a creative editing task. 
\section{Learning to Ground Language with Abstructions}
\label{sec:method}

\begin{figure*}[!tb]
    \center{\includegraphics[width=\textwidth]{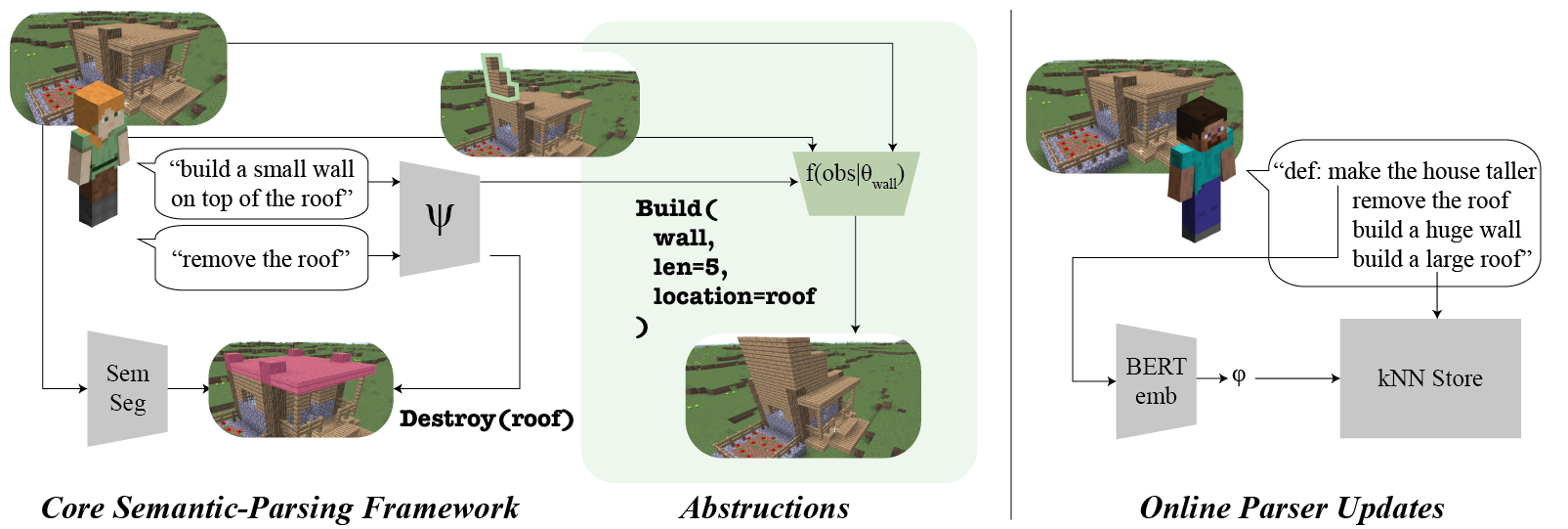}}
    \caption{\label{fig:control-flow} Overview of control flow when interacting with agent}
\end{figure*}

We study whether making abstructions, i.e., well-formed assumptions about label-conditioned generative models and their associated constraints, available to a semantic parsing framework can enable precision in specifying new commands without compromising the flexibility to work across contexts.
To this end, we focus on an open-ended, creative house modification task in Minecraft. Users are able to specify any desired modifications to a given home through natural language requests to a virtual agent. When the virtual agent does not know or understand a given request, users have the opportunity to define new requests in terms of utterances the virtual agent already understands.
\paragraph{System Overview.} At the start of interaction with users, the agent has access to a \textit{core semantic-parsing framework} (Figure~\ref{fig:control-flow}a), capable of building and destroying various house parts.
We introduce \textit{abstructions} (Figure \ref{fig:control-flow}b) into the \texttt{Build()} operation of the semantic parser by training label-conditioned generative models of next block placement and creating a set of user controls---number of blocks placed, build prompt, and location---that affect the quality and meaning of generated blocks.
As users supply the agent with new instruction-program pairs, \textit{online parser updates} (Figure \ref{fig:control-flow}c) are made through an alternative framework that relies on similarity search over sentence embeddings.

\subsection{Motivating Examples}
We motivate our design by describing two examples of user-defined instructions and how they are enabled by our system. These are also illustrated in Figure \ref{fig:pull}.

\paragraph{Instructions from Compositions.} The most common form of user-defined instruction is a composition of instructions. For example, ``make the house taller'' decomposes into the following steps: ``remove the roof'', ``build a huge wall'', and ``build a large roof''. Because we handle each of these sub-instructions with generative models, as opposed to a single reference object, we can capture the ambiguity in each instruction and generalize better across homes.
By contrast, if we were to compose two different objects defined under the framework presented by \citet{wang-etal-2017-naturalizing} or \citet{karamcheti2020learning}, those objects could not adapt to a given scene because they have a fixed reference.
Additionally, new object categories can also be created from compositions of commands. For example, to ``build a second story'' users may instruct the agent to ``remove the roof, build a huge wall on top of the house, and build a roof on top of the house''.

\paragraph{Instructions from Constraints.} The benefit of abstructions is most evident in the second form of user-provided redefinitions. By placing constraints on the inference procedure of generative models, users can synthesize new object categories without requiring training a new model.
Consider the command ``build a skylight'', which can be defined as ``build a tiny window on top of the roof''. The parser invokes the generative model for ``window'' to infer two block placements starting from a location on the roof. Another example is ``build a rooftop patio'', which could be constructed with the command ``build a fence on top of the roof''. Again, the constraints define the meaning of the primitive in context. In this case, the generative model for ``fence'' doesn't stand in for a fence or a subcategory of fence, but instead it is used for its likeness to a railing. Here, the program defines the meaning of the object and that meaning is passed directly as an instructional example to the agent.

\subsection{Core Semantic-Parsing Framework}
 In the general case, we assume that we have access to a parser over a set of core instructions as well as a semantic segmentation and a sequential generation model. We build our virtual agent on top of the CraftAssist framework \citep{gray2019craftassist,srinet-etal-2020-craftassist}. This software provides the tooling for creating Minecraft sessions and the virtual agent, including the semantic parsing system and semantic segmentation module that make up the core parsing framework. These assumptions are highly practical, even in more natural settings. For example, parser capable of decoding only the most straightforward core programs can be trained entirely through augmented data \citep{Marzoev2020Unnatural} or through a very small set of annotated utterances.

\paragraph{Core Parser.}We use the parser introduced by \citet{srinet-etal-2020-craftassist}, which directly trains a BERT-based neural semantic parser on high-level Minecraft actions. Although the parser is capable of interpreting commands about a variety of high level tasks, we focus on \texttt{Destroy()} and \texttt{Build()} actions. The \texttt{Destroy()} action operates over a fixed collection of object categories, which are segmented from the scene as described in the next paragraph. For the \texttt{Build()} operation, the parser is also capable of interpreting qualitative or quantative descriptions of length and relative location.

\paragraph{Segmentation for \texttt{Destroy()} Operations.} We use the semantic segmentation module provided by \citep{gray2019craftassist} to infer objects to destroy from the current house state. Unlike our \texttt{Build()} operation, we do not use user-specified constraints to steer the inference procedure of the \texttt{Destroy()} action, although this would be an interesting direction for future work.

\subsection{Abstructions for \texttt{Build()} Operations}
We replace the \texttt{Build()} action that is native to the CraftAssist framework with abstructions: a set of constraints on the inference procedure of a label-conditioned generative model that can affect the quality of generation and meaning of the primitive in context.

Although we describe a specific instantiation of abstructions, any generative model and set of constraints could be integrated into a semantic parser for the same purpose. In the general case, we assume that we have access to a parser, $\psi$, that infers constraints and labels, $c$, from utterances, $u$.
The inferred label indexes into a library of fixed parameters, $\theta_c$, which condition the generator. The user can also provide direct interventions to guide the generation process. Each of these components is visualized in Figure \ref{fig:control-flow}.

In our setting, we use a sequential block placement model as our generative model and a neural semantic parser trained on high-level Minecraft actions for $\psi$. However, any set of generative models that can support these constraints and that sufficiently cover the set of visual primitive concepts for a given task could be used as abstructions. \citet{bau2020units} is one compelling example in the domain of image editing. Our interventions take the form of prompts: users may optionally supply the agent with a structure to seed the generation process. Finally, we only handle constraints on location and length.

\paragraph{Label-Conditioned Generative Models.} To generate label-conditioned block placements, we adapt the VoxelCNN model presented in \citet{chen2019order}. Given a 3D patch of a scene with block type information and a global view with occupancy information, VoxelCNN predicts next block type and placement. The original model was trained to generate complete houses on a dataset of 2,500 homes. We use voxel-level semantic segmentation labels to fine-tune VoxelCNN models for the labels: balcony, bed, bookcase, ceiling, column, deck, door, fence, floor, foundation, garden, grass, ground, ladder, lights, patio, pillar, porch, railing, roof, stair, torch, walkway, wall, window, and yard.

Our fine-tuning runs for an additional 4 epochs and for each class we select the model with the best performance on the validation set. We use the same train-val split as \citet{chen2019order}, but save 50\% of the validation set as our test set. Averaging across categories, we achieve a top-10 accuracy of 66.0\% and average 7.50 consecutively correct blocks. Performance by category as well as hyperparameters and compute resources can be found in Section A.1. For the classes window and bed, we overwrite block type prediction so that the agent gives predictable results. Window is hard-coded to predict glass blocks and bed is hard-coded to predict bed blocks.

For our setting, a sequential model is a very beneficial design choice. Users are able to provide interventions that strongly influence the outputs of the model so that concepts can be reused or remixed. Users can also prevent compounding errors by calling the same generative model multiple times. For example, if the direction of a wall is ambiguous, a user can instruct the \texttt{Build()} action to be called several times, allowing for them to intervene upon error. Our selection of constraints are motivated by these intervention opportunities.

\paragraph{Constraints and User Controls.} We allow users to steer the inference procedure by providing constraints on location and length and intervening to supply a prompt. Location and length can be specified through natural language. If no location is specified, the agent asks users to specify a hint for the generation process. At this time, the user can provide location and prompt suggestions.
\begin{itemize}
    \item \textit{Location} is the coordinate at which the generative procedure will start. If a user specifies a location in their instruction, such as "on top of the roof", the coordinate location will be resolved heuristically using tools from the original CraftAssist framework. Otherwise, the starting coordinate is the direction of the user's cursor is projected onto the nearest house block.
    \item \textit{Length} is the number of voxels sampled from the generative model. We map qualitative descriptions of size to block types: tiny is 2 blocks, small is 5 blocks, large is 50 blocks, huge is 100, and the default is 20 blocks. Note that this mapping is limited as qualitative descriptions of size don't adapt to category or to house size. It does however give users much more predictable results when specifying commands.
    \item \textit{Prompt} is a block structure that users can optionally provide when asked for a hint. Depending on the primitive that is being invoked, this could affect the block type predicted or the final structure shape. The system does not memorize or retain information about prompts, but this would be an interesting direction for future work.
\end{itemize}

\subsection{Parser Updates Online} Users are able to define new commands with the following syntax: 
\begin{center}
  \texttt{def: new command; sub command 1; \ldots; sub command N}  
\end{center}
Below we denote the \texttt{new command} as $u$. Upon receipt of a user-provided definition, we retrieve embeddings for each token of $u$. Denoting $m$ as the sentence length, we have:
\begin{align}
    \phi_{1:m} &= \text{BERT}(u_{1:m})
\end{align}
Our BERT embeddings are provided by the HuggingFace Transformers \cite{wolf-etal-2020-transformers} library. To achieve a single sentence representation, $\phi$, we apply an aggregation function across the token features:
\begin{align}
    \phi &= \sigma(\phi_{1:m})
\end{align}
We use averaging as our choice for $\sigma$. This method draws inspiration from \citet{karamcheti2020learning}, but does not utilize additions like added aggregation layers or lifted utterances as in our setting, changing a reference object can substantially change the inferred program.

When new requests are received, the aggregated features of the request are cross referenced with a nearest neighbor store before a parse is attempted. We use Facebook AI Similarity Search \citep{JDH17} as our nearest neighbor store. For the scale of data we collect in our experiments, training a layer on top of features for improved embedding quality is ineffective, so this essentially acts as a dictionary lookup.
\section{Experiments}
\label{sec:exp}



\begin{figure}[!t]
    \center{\includegraphics[width=\textwidth]{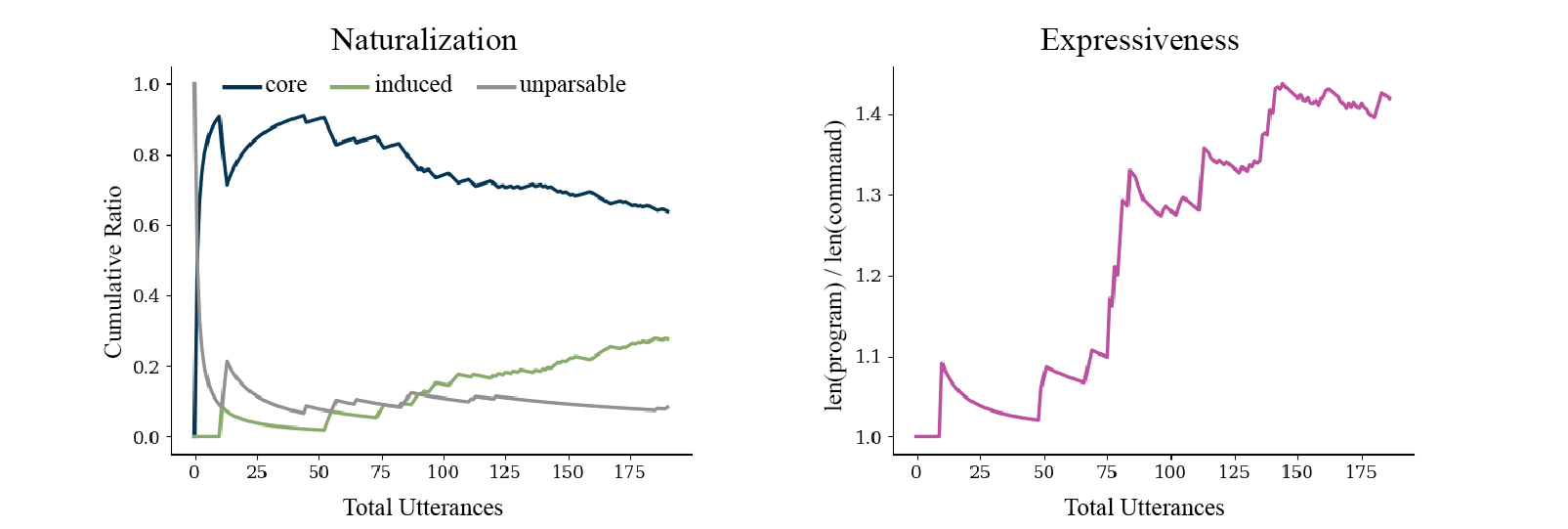}}
    \caption{\label{fig:exp} \textit{Left}. Expressiveness is measured as the sum of core commands (i.e., build or destroy a reference object) and constraints each utterance refers to. \textit{Right}. As users define more commands, users rely increasingly on induced utterances to achieve their build goals.} 
\end{figure}


We show that over time, users rely increasingly on induced utterances and that the average utterance becomes more expressive. We also evaluate the generalization performance of user-defined commands with user surveys and qualitative examples.

\subsection{Experiment Design}
\label{sec:exp_design}
Our experiment ran in two phases: one qualifying task that introduced users to the agent's capabilities and limitations, and a creative build period where users could define new instructions, shared across the entire user population. All of the data we present is sourced from the creative build period, which was designed as follows. In each session, users enter a Minecraft server with the agent three separate times and interact with a total of two different homes. In the first instance, the users have the opportunity to test out any modification requests or new commands on the first home. In the second instance, the users apply their desired modifications and new command definitions to the same home. All of these modifications are specified through dialogue exchanges with the agent. Users are also welcome to ``clean up'' the agent's work, but many users chose not to clean up the agent's block placements. By creating two separate sessions with the same house, we allow users to explore the capabilities and weaknesses of the bot. In the third session, we instruct users to replicate all of their modifications on a new, second home so that we can evaluate how well their re-definitions transfer. 

During the creative build period, we explicitly told users that our goal was to teach the agent new, expressive instructions. We also provided videos describing tips to get good results out of the generative models and ideas about types of commands to define. The full set of instructions given to users in both the qualifying task and creative build period are provided in the appendix. We pre-populate our nearest-neighbor store with the commands defined in the qualifying task and in the tutorial videos. This includes ``make the house taller'', ``build a skylight'', and ``make me a place to sit down''. Each house is randomly sampled from the test split we use for training the label-conditioned generative models. We filter house candidates from the test split based on dimension in voxels, so that the size terms apply well across different homes. We also remove homes with blocks that behave atypically, such as lava or water. This leaves us with 23 total homes from which to sample. 

We hosted our experiments on Amazon Mechanical Turk and handled HIT management through the EasyTurk wrapper \citep{krishna2019easyturk}. Compute details for the AMT experiments are provided in Section A.2. Of the 11 users who attempted the qualifying task, 8 passed and 4 chose to continue with the creative task. All users were paid twelve dollars per hour, with bonuses reaching fifteen dollars per hour. Prior to launching the the Minecraft server, users were notified that the data from their interactions will be collected and were instructed not to share any personally identifying information. We also removed their Minecraft username from the data we provide. 

\subsection{Naturalization}
To evaluate whether naturalization still takes place when we introduce abstructions, we classify every dialogue exchange that results in a \texttt{Build()} or \texttt{Destroy()} action (i.e., not conversational exchanges like ``hello'') as `core', `induced', or `unparsable'. Core utterances are actions that the agent could complete successfully with the core parser alone. Induced utterances are commands that were defined by the user population. Unparsable utterances are commands that the agent could not complete successfully. Examples include syntax or reference objects with which the agent is unfamiliar. The agent can also fail to parse successfully because of computational issues unrelated to its abilities, such as the segmentation model not inferring objects from the scene quickly enough. These are still counted as unparsable. We only consider dialogue exchanges from the second and third sessions of the creative build task. Note, we explicitly ask the users to test their redefinitions in a new environment. However, we see a similar result for dialogue exchanges in the second session only. Please see Section A.3 of the appendix for these results. We also treat new command definitions at the time that they're provided as the sequence of commands that define it. For example, we count ``def: make the house taller; remove the roof; build a huge wall; build a large roof'' as three core utterances, as opposed to one induced utterance. 

Figure \ref{fig:exp} shows the cumulative ratio of parsable, unparsable, and induced utterances over the course of all user dialogue exchanges. Similar to \citet{wang-etal-2017-naturalizing}, we see a steady increase in the proportion of utterances that come from user-induced commands. However, we do not see the proportion of induced utterances overtake core utterances. We suspect that this is because the core actions we provide cover more modifications of interest within our setting. Unlike in \citet{wang-etal-2017-naturalizing}, the users do not need to start from single block placements. Over the entire course of naturalization, 27.7\% of utterances were induced.

\subsection{Expressiveness} We define expressiveness as the length of the ``program'', i.e., user-provided redefinitons, divided by the length of the utterance that maps to it. For a core utterance, the expressiveness is one because the ``program'' is simply the original command. The expressiveness of an induced command is the total number of words or tokens in the commands that are specified in the definition. For example, ``build a skylight'' has 3 words and is defined as `build a tiny window on the roof'', which has 7 words. So, the expressiveness of ``build a skylight'' is 2.33. As with naturalization, we compute expressiveness for the second and third sessions of the creative build task. We again treat new command definitions as the sequence of commands in the definition.

Figure \ref{fig:exp} shows that expressiveness increases over the course of the experiment, achieving an average expressiveness of 1.42 by the end of our naturalization experiment. Some users define commands with expressiveness below one. For example, one user defined the command ``build an awning on house'' as ``build a roof'', leading to an expressiveness score of .6. Presumably these redefinitions still have value to users or are otherwise more natural because of the prompts they are able to provide. Cases such as this could explain the dips in expressiveness across training. It also indicates that this way of computing expressiveness, does not fully capture or describe the benefits that redefinitions provide to users.

\begin{figure}[!t]
    \center{\includegraphics[width=\textwidth]{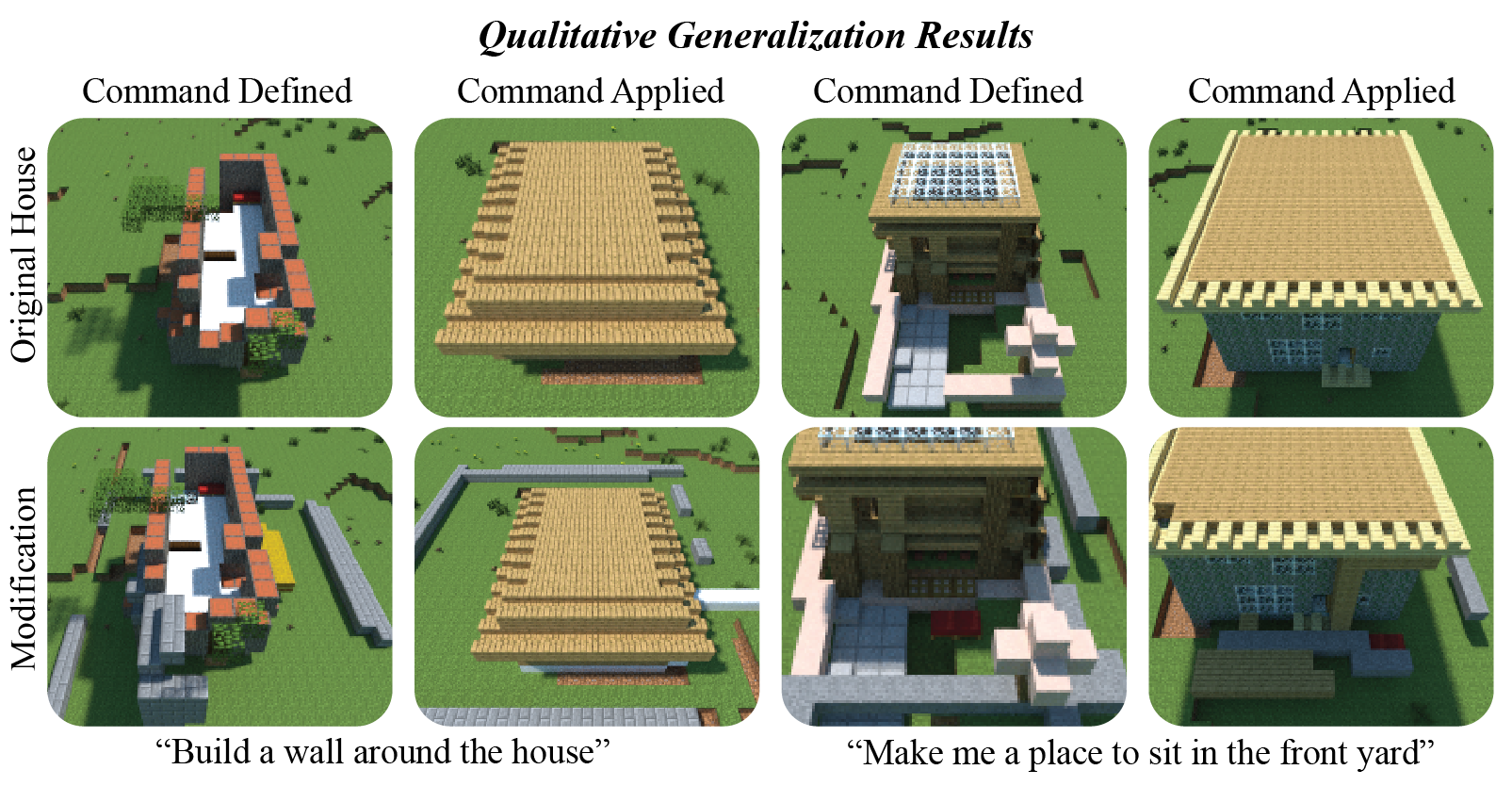}}
    \caption{\label{fig:generalization} Examples of home modifications generated from user commands. The top row shows the original, unmodified home and the bottom row shows the result after the modification is applied. The first and third columns are examples where the user defined the command. The second and fourth column show the command applied to a new house.} 
\end{figure}

\subsection{Generalization across Diverse Homes} 

We use user surveys and qualitative examples of command transfers to understand how well abstructions enable redefinitions across the second and third session of the creative build task. Upon completing the task, 100\% of users agreed that the command successfully transferred. Qualitative results revealed strong variance in the performance of generalized commands. The discrepancy between user ratings of generalization performance and visualizations of transfer performance are likely due to ambiguity in the definition of a successful transfer.

In Figure \ref{fig:generalization}, we show how well two commands generalize across homes. The command ``build a wall around the house'' is defined as ``build a wall; build a wall; build a wall; build a wall''. The agent successfully builds walls around both homes. The command ``make me a place to sit in the front yard'' is defined as ``make me a place to sit in the front yard: make a fence around the house; make me a place to sit down''. Both of these are themselves induced commands. In this case, the agent makes a somewhat enclosed space and places two bed blocks, which are the small red blocks in front of the home, as a place to sit down.

\subsection{Examples of Commands Defined}
\label{sec:command_eg}
\begin{figure}[t]
    \center{\includegraphics[width=\textwidth]{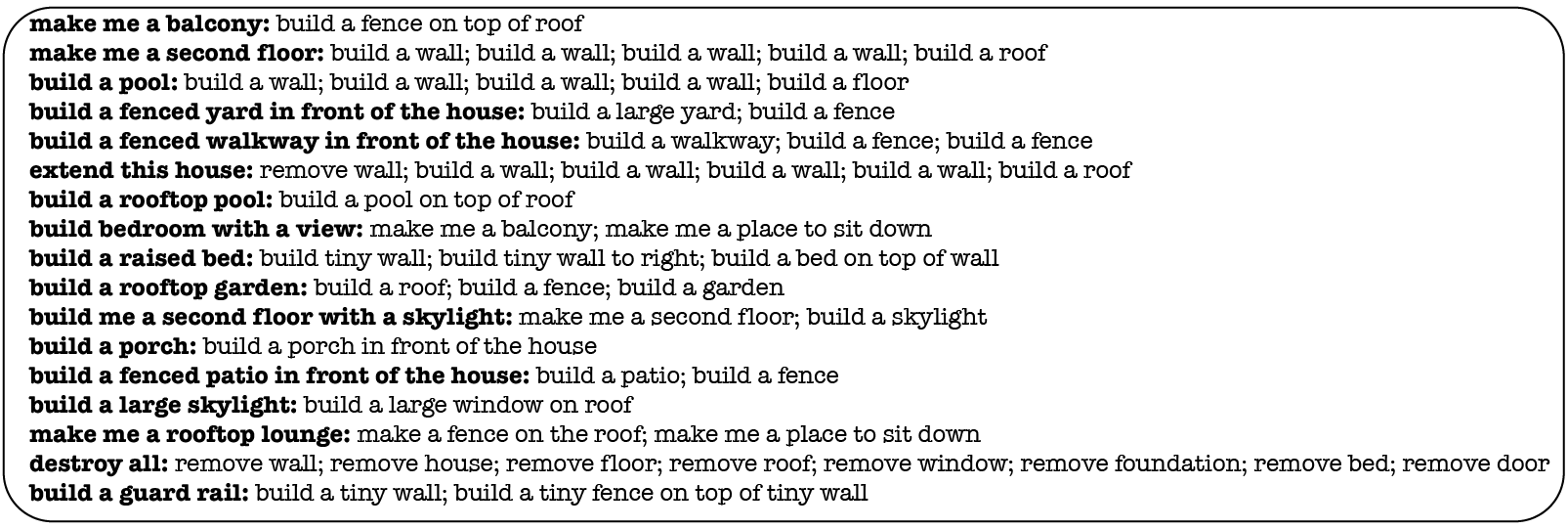}}
    \caption{\label{fig:commands} Examples of user-provided redefinitions.} 
\end{figure}
We visualize a selection of the 38 total commands supplied by users in Figure \ref{fig:commands}. These show that users significantly borrowed from other user's commands. These examples also illustrate some weaknesses of our design. Not all commands generalize to different forms of homes; ``destroy all'' only works on homes with that exact list of objects. Users also relied heavily on the ability to specify location hints from their cursor, which is why many commands don't include qualitative descriptions of location.
\section{Conclusion}
\label{sec:conc}
In this work, we present the idea of neural \textit{abstructions}: a set of abstractions around the generative models and associated constraints that make it easier for users to develop precise commands that generalize across contexts. Our results show that abstructions have the potential to bring naturalization frameworks to a broader set of creative build tasks than was initially shown in \citet{wang-etal-2017-naturalizing}. There are several directions we recommend for future work to realize this goal in more realistic setting: building sets of generative models for natural images that could be steered and repurposed in the same way that our block-placement models can and developing algorithms that better take advantage of user-specified programs to make the benefits of data collection even more fruitful.

\section*{Acknowledgements}
We thank Siddharth Karamcheti, Ranjay Krishna, Shyamal Buch, and Eric Mitchell for helpful discussions and feedback. Thank you to Alex Tamkin for coming up with the name ``abstructions''. Kaylee Burns is graciously supported by the NSF graduate fellowship. Christopher D. Manning is a CIFAR Fellow.

\bibliographystyle{plainnat}
\bibliography{anthology,neurips_2021}

\appendix

\section{Appendix}
\subsection{Generative Model Details}
All models are trained on a TeslaK40c through an internal cluster. We compared models across the following hyperparameters: with and without block embeddings, on a feature dimension size of 16, 32, and 64, on placement histories of size 1 and 3, with learning rates of .1 and .01, and on batch sizes of 32 and 64. Our final model uses block embeddings and has a feature dimension size of 32, history length of 3, learning rate of .1, and batch size of 64.

Our model's block placement prediction performance across categories is depicted in Table \ref{tab:perf}. Accuracy at 10 denotes the accuracy of block placements within the first 10 predictions of the model. CCA Average is the average consecutively correct block placements averaged across home completion amounts of 10\%, 25\%, 50\%, 75\%, and 90\%. These metrics are adopted from \citet{chen2019order}.

\begin{table}[!h]
  \caption{Fine-tuned VoxelCNN performance across object categories}
  \label{tab:perf}
  \centering
  \begin{tabular}{lll}
    \toprule
    \cmidrule(r){1-2}
    label & Acc@10 & CCA Avg. \\
    \midrule
    floor & 0.83 &11.28  \\
    roof & 0.84 & 10.82 \\
    foundation & 0.76 & 10.41\\
    wall & 0.78 & 11.02\\
    walkway & 0.66 &9.56 \\
    ceiling & 0.81 & 10.07 \\
    balcony & 0.73 &8.86 \\
    stairs & 0.41 &8.36 \\
    patio & 0.79 & 8.10\\
    porch & 0.71 & 8.31\\
    deck & 0.71 &8.31 \\
    pillar & 0.65 &7.86 \\
    window & 0.84 &8.08 \\
    lights & 0.42 &7.31 \\
    column & 0.59 &7.17 \\
    door & 0.51 &6.29 \\
    ground & 0.64 &7.25 \\
    torch & 0.30 & 6.79\\
    railing & 0.79 & 5.56\\
    fence & 0.79 & 5.45\\
    grass & 0.62 & 4.62\\
    bookcase & 0.63 &4.07 \\
    garden & 0.51 & 3.43\\
    yard & 0.51 &0.95 \\

    \bottomrule
  \end{tabular}
\end{table}

\subsection{Naturalization experiment}
\begin{figure}[!h]
    \center{\includegraphics[width=\textwidth]{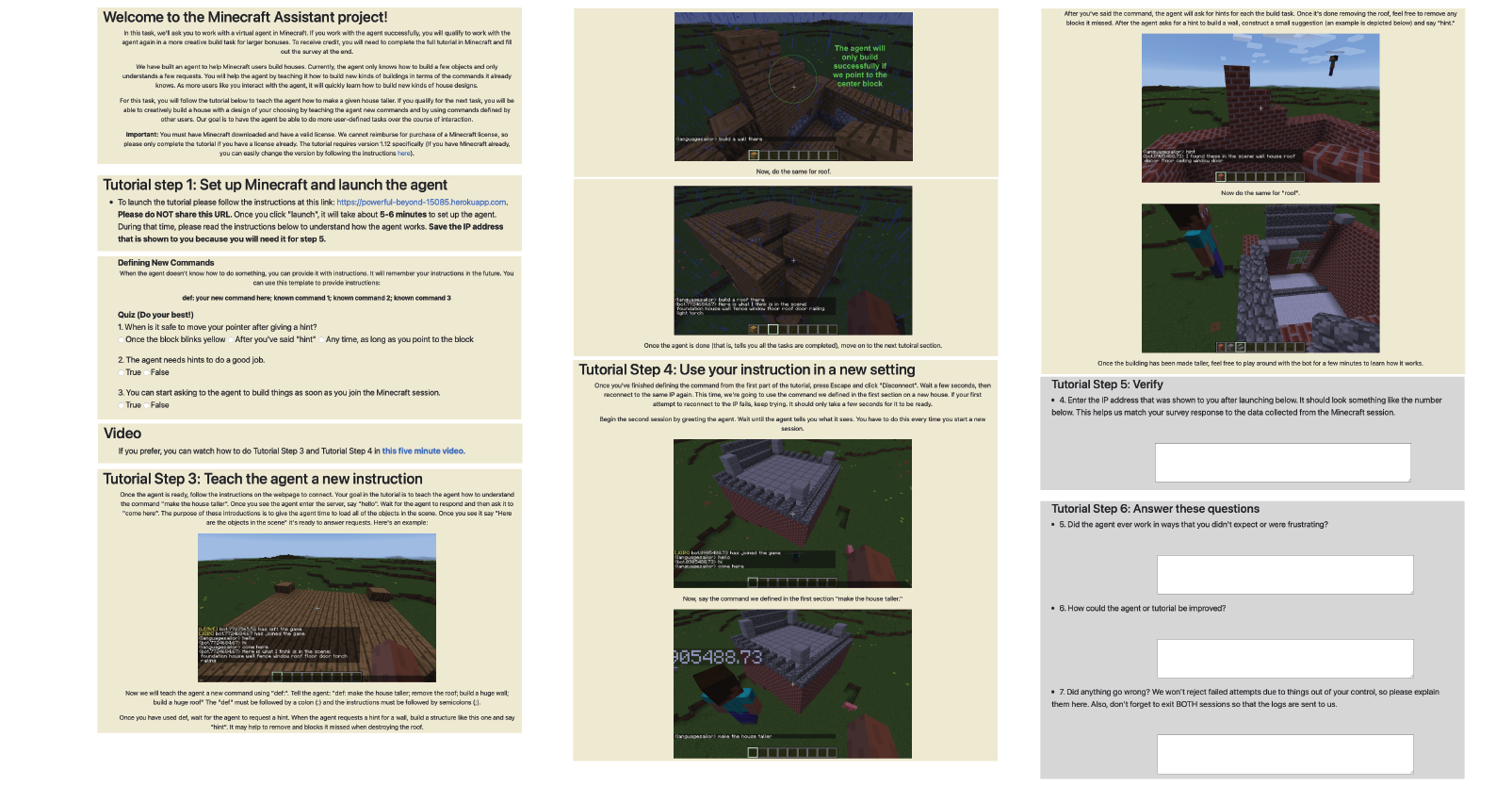}}
    \caption{AMT qualification task description.}
    \label{fig:mturk_qual}
\end{figure}

\begin{figure}[!h]
    \center{\includegraphics[width=\textwidth]{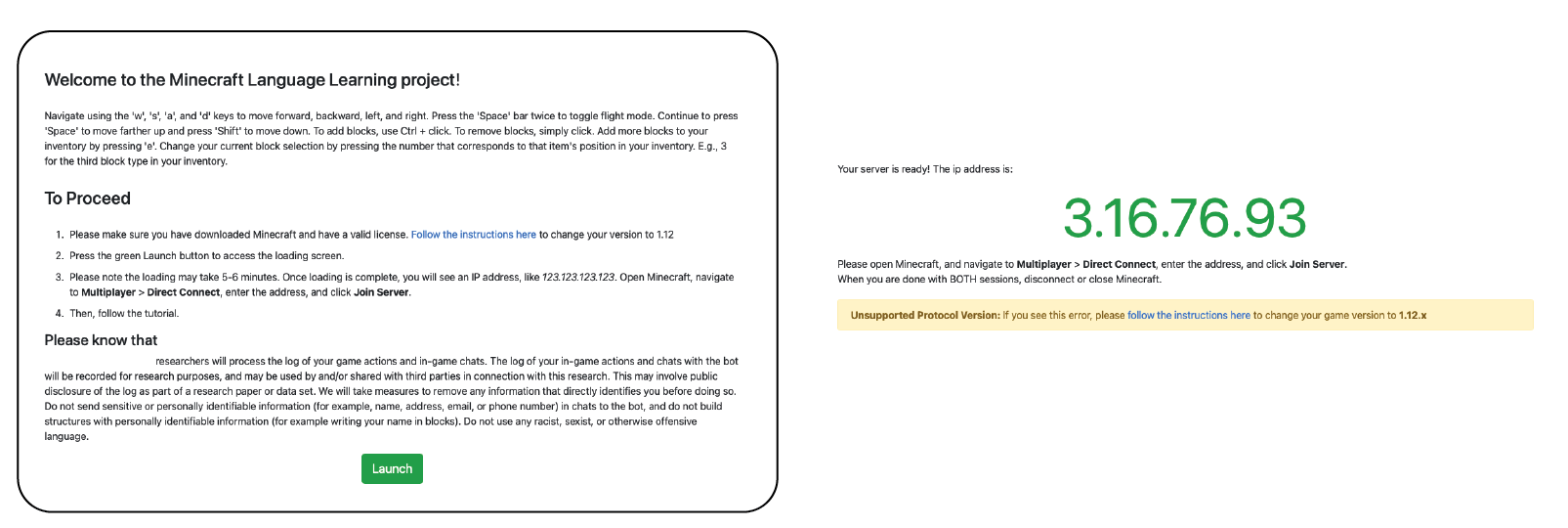}}
    \caption{Server website for qualification task.}
    \label{fig:serv_qual}
\end{figure}

\begin{figure}[!h]
    \center{\includegraphics[width=\textwidth]{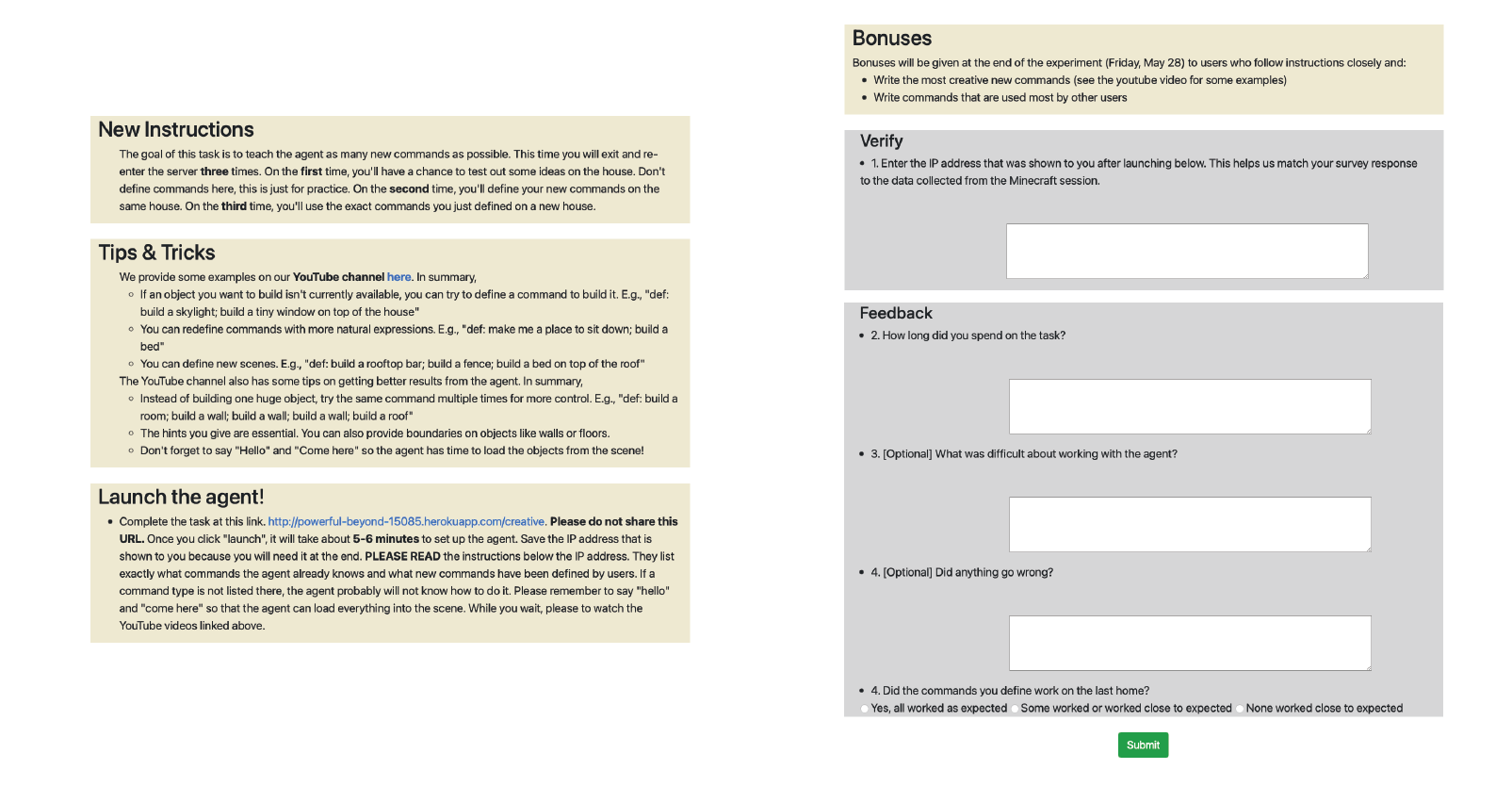}}
    \caption{AMT main experiment task description.}
    \label{fig:mturk_exp}
\end{figure}

\begin{figure}[!h]
    \center{\includegraphics[width=\textwidth]{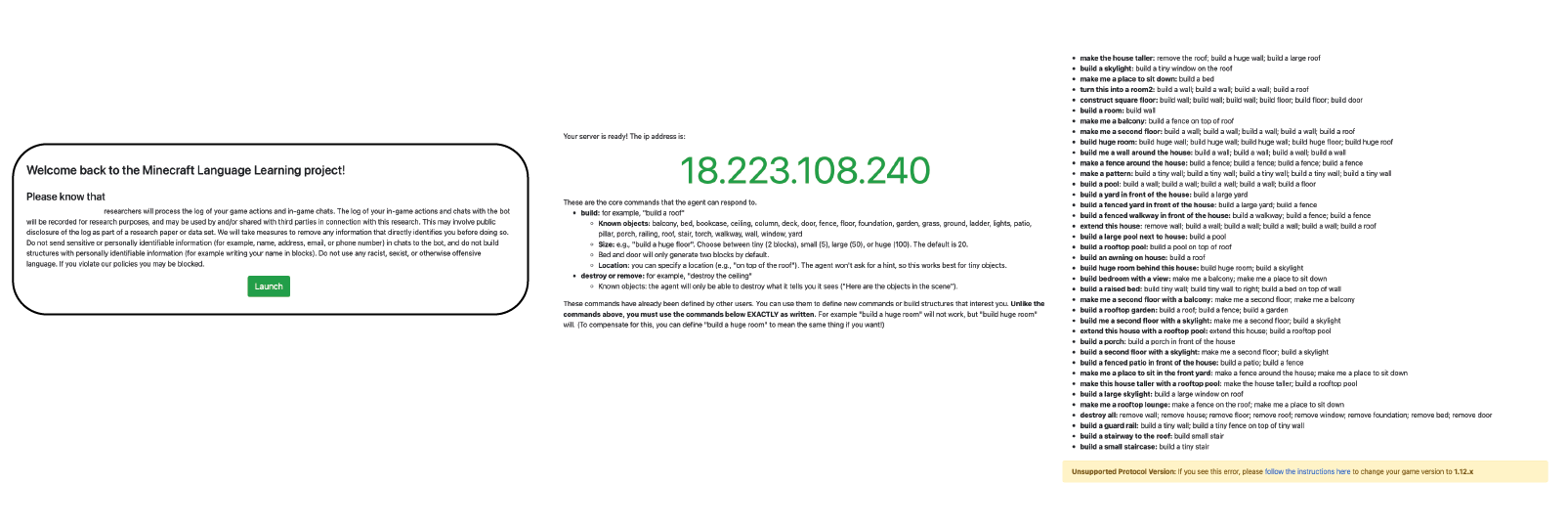}}
    \caption{Server website for main experiment.}
    \label{fig:serv_exp}
\end{figure}

Our data collection occurred in two stages: we hosted a qualifying task, during which users were instructed to follow a tutorial video to familiarize themselves with the agent, and the main experiment, which was an open ended house modification task. For each task, we walked users through instructions on Amazon Mechanical Turk (AMT)  and then directed them to a website, which launched a Minecraft server for them to connect to. The AMT instructions and server instructions for the qualifier are in Figures \ref{fig:mturk_qual} and \ref{fig:serv_qual}. The AMT instructions and server instructions for the main experiment are in Figures \ref{fig:mturk_exp} and \ref{fig:serv_exp}.

We hosted the Minecraft server and agent on ECS instances. A new task is run on each launch with a memory size of 8192 MiB and 4096 CPU units.


\subsection{Naturalization and Expressiveness}
\begin{figure}
    \center{\includegraphics[width=\textwidth]{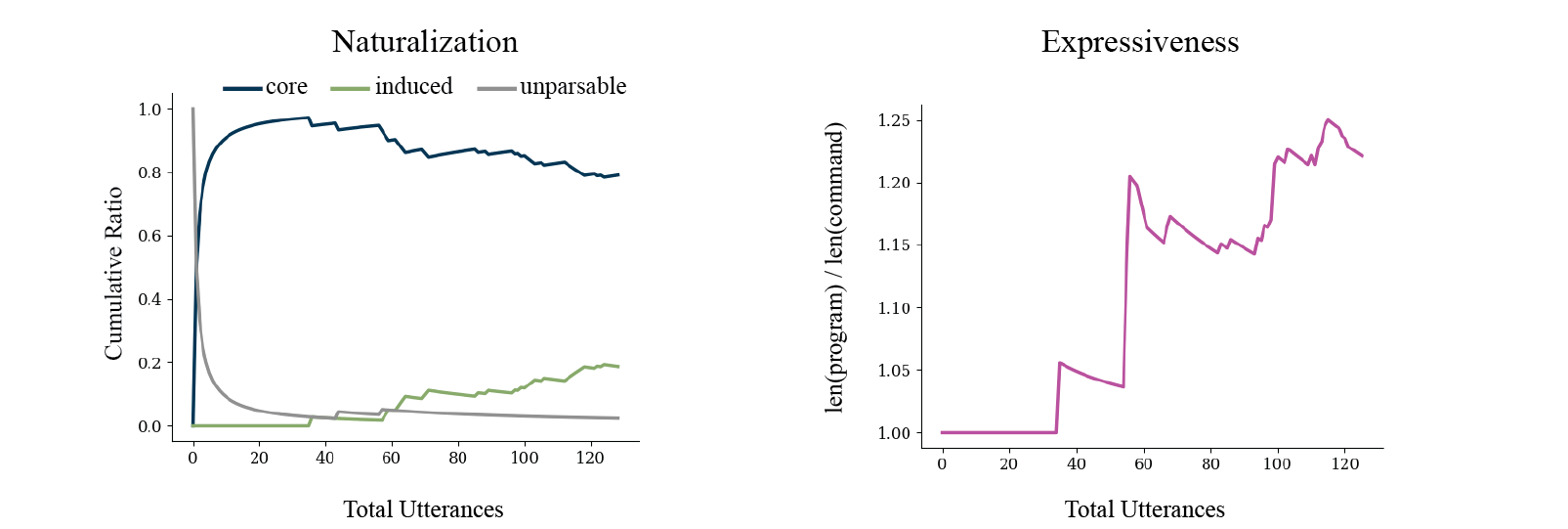}}
    \caption{Naturalization and expressiveness results over only the second session, where users are not explicitly asked to repeat defined commands. When classifying utterance type, we use each sub-command in a command definition.}
    \label{fig:exp_2}
\end{figure}

To verify that the induced commands were not limited to the third session, where users are explicitly asked to repeat defined commands, we plot naturalization and expressiveness results from just the second session in Figure \ref{fig:exp_2}.

\end{document}